\documentclass[10pt,letterpaper,twocolumn]{article}

\usepackage[margin=1.9cm]{geometry}

\usepackage{hyperref}
\usepackage{graphicx}
\usepackage{caption}
\usepackage{subcaption}

\begin{document}

\twocolumn[

\title{The Collective Knowledge project: making ML models more portable and reproducible with open APIs, reusable best practices and MLOps}

\author{Grigori Fursin \\
\\
cTuning foundation and cKnowledge SAS
\\
\\
\href{https://cKnowledge.io}{cKnowledge.io}\\
}

\vskip 0.2in

\maketitle

\begin{abstract}

This article provides an overview of the Collective Knowledge technology (CK or cKnowledge).
CK attempts to make it easier to reproduce ML\&systems research, deploy ML models in production,
and adapt them to continuously changing data sets, models, 
research techniques, software, and hardware.
The CK concept is to decompose complex systems and ad-hoc research projects
into reusable sub-components with unified APIs, CLI, and JSON meta description.
Such components can be connected into portable workflows using DevOps principles 
combined with reusable automation actions, software detection plugins, meta packages,
and exposed optimization parameters.
CK workflows can automatically plug in different models, data and tools from different vendors
while building, running and benchmarking research code in a unified way across diverse platforms and environments.
Such workflows also help to perform whole system optimization, reproduce results, and compare them 
using public or private scoreboards on the \href{https://cKnowledge.io}{cKnowledge.io} platform.
For example, the modular CK approach was successfully validated with industrial partners
to automatically co-design and optimize software, hardware, and machine learning models for reproducible
and efficient object detection in terms of speed, accuracy, energy, size, and 
other characteristics.
The long-term goal is to simplify and accelerate the development and deployment of ML models
and systems by helping researchers and practitioners to share and reuse their 
knowledge, experience, best practices, artifacts, and techniques using open CK APIs.

\end{abstract}

\vspace{0.3cm}


\vspace{0.3cm}

]

\section{Motivation}

\begin{figure*}[ht]
  \centering
  \includegraphics[width=1.0\textwidth]{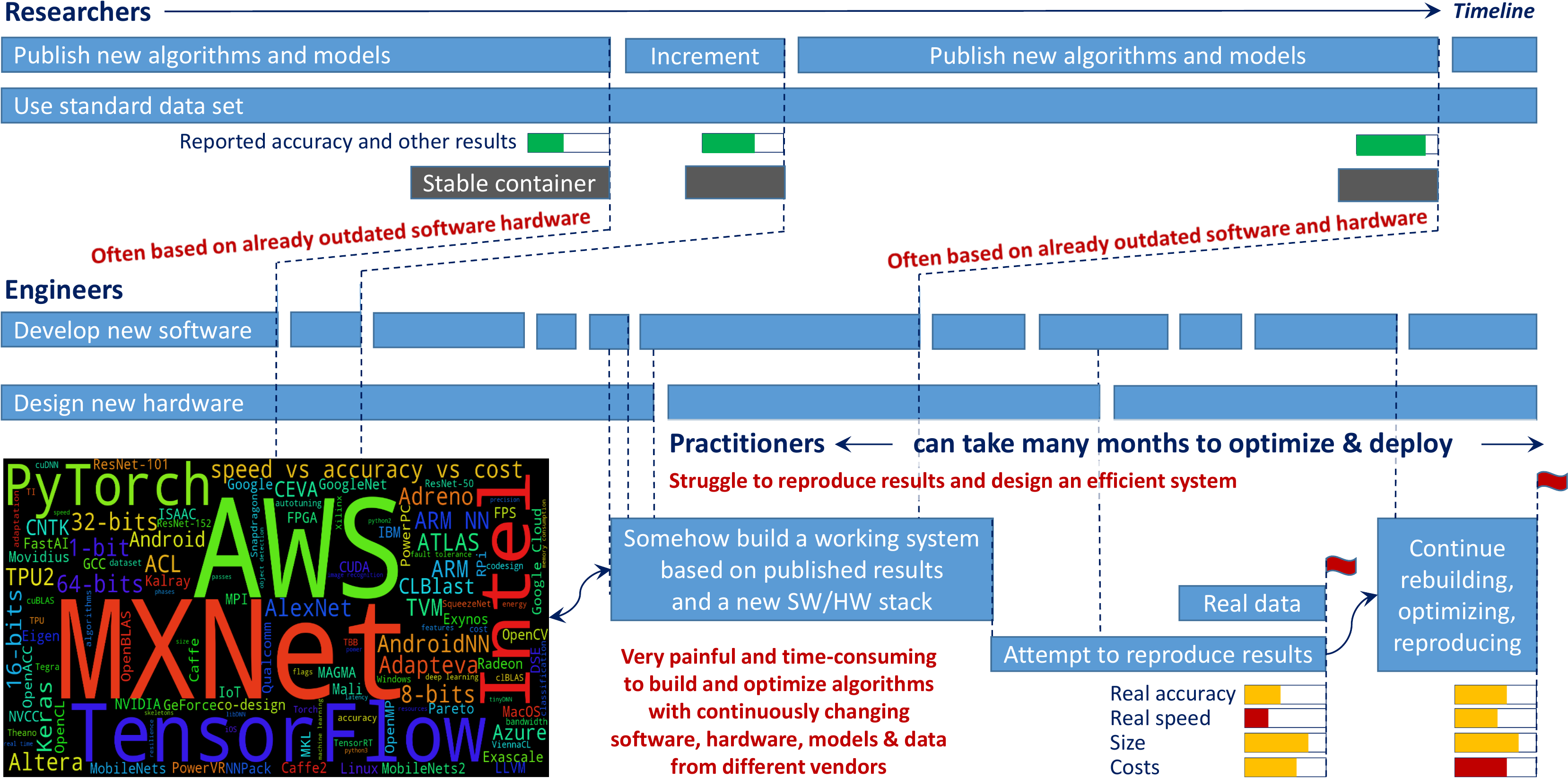}
  \caption{Reproducing research papers and adopting novel techniques in production 
   is a tedious, repetitive and time consuming process
   because of continuously changing software, hardware, models and datasets, 
   and a lack of common formats and APIs for shared code, models, and artifacts.}
  \label{fig:timeline1}
\end{figure*}

10 years ago I developed the \href{https://cTuning.org}{cTuning.org} platform and released all my research code and data 
to the public to crowdsource the training of our machine learning based compiler (MILEPOST GCC)~\cite{Fur2009}.
I intended to accelerate this very time consuming autotuning process and help our compiler to learn the most efficient optimizations 
across real programs, data sets, platforms, and environments provided by volunteers. 

We had a great response from the community and it took me just a few days to collect as many optimization results 
as during my entire PhD research.
However, the initial excitement quickly faded when I struggled to reproduce most of the performance numbers and ML model predictions
because even a tiny change in software, hardware, environment and the run-time state of the system could influence performance
while I did not have a mechanism to detect such changes~\cite{mdhs2009,DBLP:journals/corr/FursinD14}.
Even worse, I could not compare these empirical results with other published techniques because they rarely included
the full experiment specification and all the necessary artifacts along with shared research code to be able to reproduce results.
Furthermore, it was always a nightmare to add new tools, benchmarks and data sets to any research code
because it required numerous changes in different ad-hoc scripts, repetitive recompilation of the whole project 
when new software was released, complex updates of database tables with results, and so on.

These problems motivated me to establish the non-profit cTuning foundation and work on a common methodology and open-source tools
to enable collaborative, reproducible, reusable, and trustable R\&D.
My foundation has supported multiple reproducibility initiatives 
at systems and machine learning conferences in collaboration with ACM.
We also promoted sharing of code, artifacts
and results in a unified way along with research papers~\cite{AE,cfkz2016}.
It gave me a unique chance to participate in reproducibility studies of more than 100 research papers 
at MLSys, ASPLOS, CGO, PPoPP, Supercomputing, and other computer science conferences during the past 5 years~\cite{ae-reproduced-papers}.
I also started deploying some of these techniques in production in collaboration with my industrial partners 
to better understand all the problems when building trustable, reproducible, and production-ready computational systems.

This practical experience confirmed my previous findings: while sharing ad-hoc research code, artifacts, and trained models
along with research papers is a great step forward, it is only a tip of the reproducibility iceberg~\cite{grigori_fursin_2017_2544204}.
The major challenge afterwards is to figure out how to integrate such
code and models with complex production systems and run them in a reliable
and efficient way across rapidly evolving software, heterogeneous hardware
and legacy platforms with continuously changing interfaces and data
formats while balancing multiple characteristics including speed, latency,
accuracy, memory size, power consumption, reliability, and costs
(Figure~\ref{fig:timeline1}).

\section{Collective Knowledge framework}

\begin{figure*}[ht]
  \centering
  \includegraphics[width=1.0\textwidth]{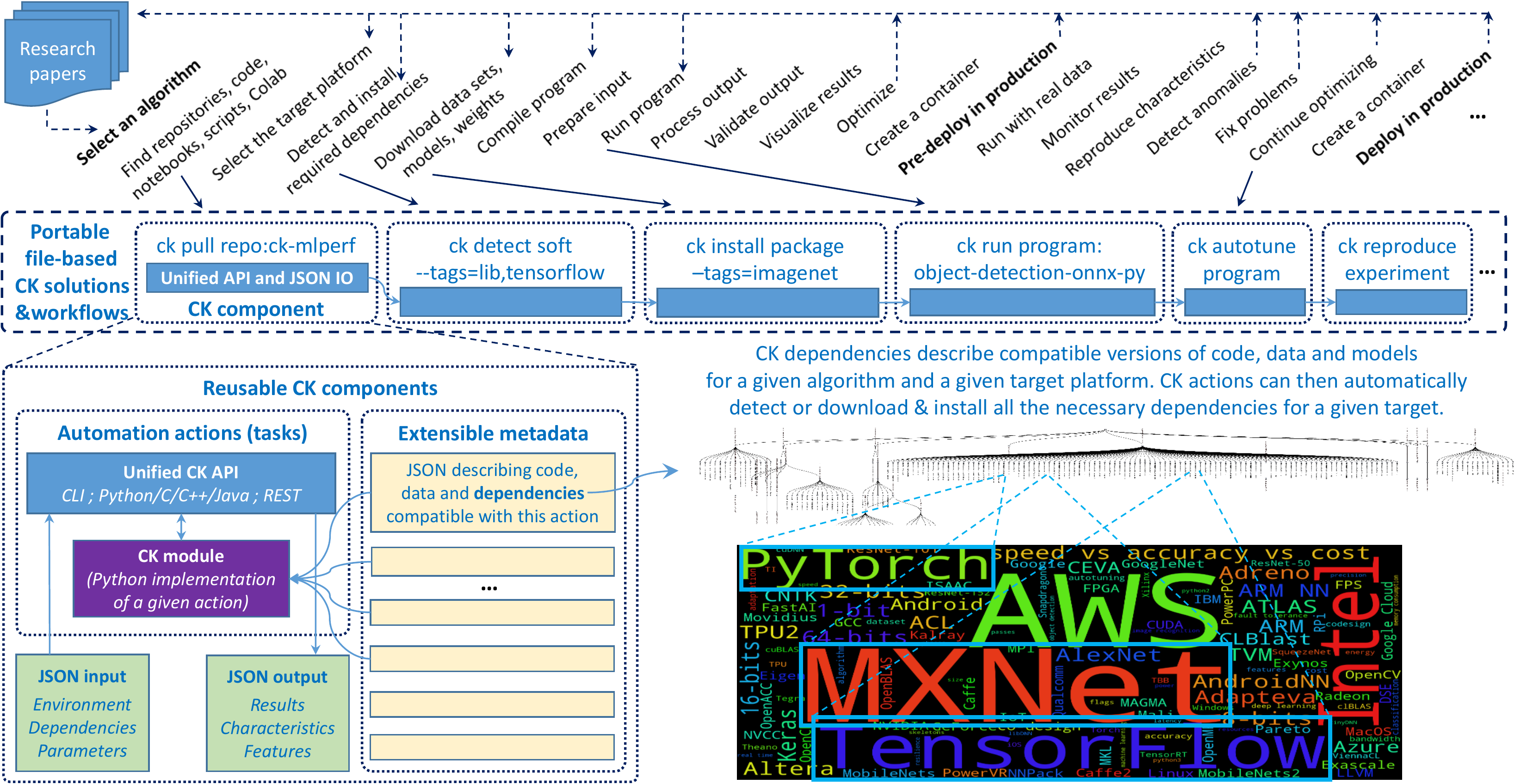}
  \caption{Collective Knowledge framework helps to convert ad-hoc research artifacts 
   (code, data, models, results) into reusable components, automation actions 
   and portable workflows with a unified CLI, Python API and JSON meta description shared along with papers. 
   The goal is to make it easier to reproduce, reuse, adopt and build upon ML\&systems research.}
  \label{fig:ck}
\end{figure*}

\begin{figure*}[ht]
  \centering
  \includegraphics[width=1.0\textwidth]{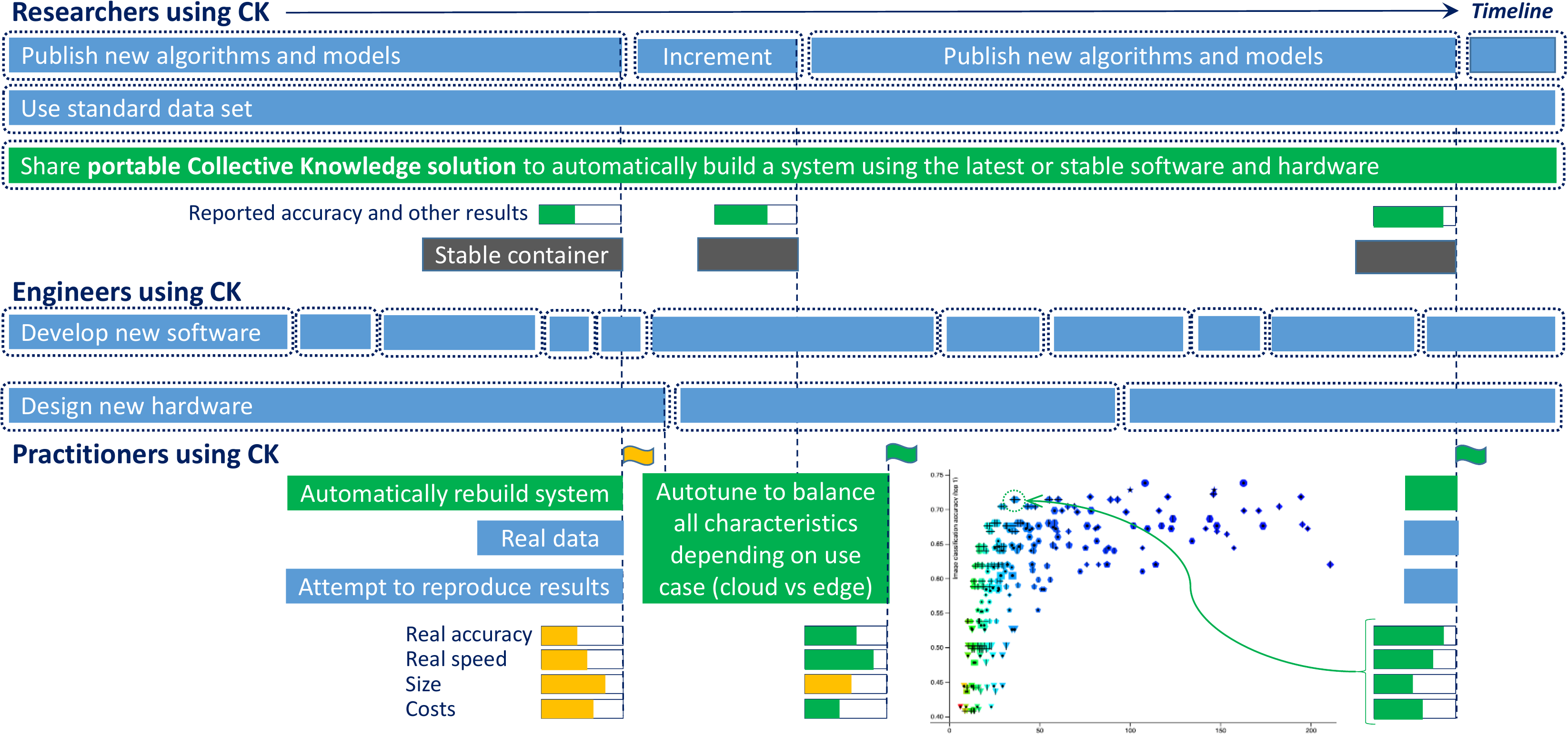}
  \caption{Portable CK workflows with reusable components can connect researchers and practitioners
  to co-design complex computational systems using DevOps principles 
  while adapting to continuously changing software, hardware, models and data sets.
  CK framework also helps to unify, automate and crowdsource 
  the benchmarking and autotuning process across diverse components
  from different vendors to automatically find the most efficient systems 
  on the Pareto frontier.
  }
  \label{fig:timeline2}
\end{figure*}

As the first step to deal with this chaos, I introduced an Artifact Appendix and a reproducibility checklist.
My goal was to help researchers describe how to reproduce their research techniques
in a unified way across different conferences and journals~\cite{ck-ae-appendix,ae-reproduced-papers}.
It was striking to notice that most of the research projects used some ad-hoc scripts often with hardwired paths 
to perform the same repetitive tasks including downloading models and data sets, detecting required software, 
building and testing research code, preparing target platforms, running experiments, 
validating outputs, reproducing results, plotting graphs, and generating papers.
This motivated me to search for a solution to automate such common tasks and make them reusable
and customizable across different research projects. 

First, I started looking at related tools that were introduced to automate experiments, 
make research more reproducible and make it easier to deploy machine learning in production:

\begin{itemize} 

\item
 ML workflow frameworks such as MLFlow~\cite{Zaharia2018AcceleratingTM}, 
 Kedro~\cite{kedro} and Amazon SageMaker~\cite{sagemaker} help to abstract and automate 
 high-level ML operations. However, unless used inside AWS or DataBricks cloud they still have limited support for the complex
 system integration and optimization particularly when targeting embedded devices and IoT - the last mile of MLOps.
 
\item
 ML benchmarking initiatives such as MLPerf~\cite{reddi2019mlperf}, 
 MLModelScope~\cite{li2019acrossstack} and Deep500~\cite{deep500}
 attempt to standardize benchmarking and co-design of models and systems.
 However, production deployment, integration with complex systems and adaptation to continuously 
 changing tools, user environments and data are out of their scope.
 
\item
 Package managers such as Spack~\cite{spack} and EasyBuild~\cite{easybuild}
 are very useful to rebuild the whole environment with fixed software versions.
 However adaptation to existing environments, native cross-compilation 
 and support for non-software packages (models, data sets) is still in progress.

\item
 Docker, Kubernetes and other container-based technology is very useful
 to prepare and share stable software releases. However, it hides all the
 software chaos rather than solving it, has some performance overheads,
 requires an enormous amount of space, have very poor support for embedded
 devices and do not help to integrate models to existing projects and
 user data.

\item
 PapersWithCode.com platform helps to find relevant research code
 for published machine learning papers and keep track of the state-of-the-art 
 machine learning research using public scoreboards. 
 However, my experience suggests that sharing ad-hoc research code 
 is not enough to make research techniques reproducible, customizable,
 portable and trustable.
  
\end{itemize} 

While working with these useful tools and platforms I realized that a higher-level
API can help to connect them into portable workflows 
with reusable artifacts that can adapt to never-ending changes 
in systems and environments.
That's why I decided to develop the Collective Knowledge framework 
(CK or cKnowledge)~\cite{ck-date16,ck2} - a small and cross-platform Python framework 
that helps to convert ad-hoc research projects into a file-based database
of reusable CK components~\cite{ck-modules} (code, data, models,
pre-/post-processing scripts, experimental results, R\&D automation
actions~\cite{ck-actions}, best research practices to reproduce results,
and live papers) with unified Python and REST APIs, common command-line interface, 
JSON meta information and JSON input/output (Figure~\ref{fig:ck}).
I also provided reusable API to automatically detect different software,
models and datasets on a user machine or install/cross-compile the missing
ones while supporting different operating systems (Linux, Windows, MacOS,
Android) and hardware (Nvidia, Arm, Intel, AMD ...).

Such an approach allows researchers to create, share and reuse flexible APIs with
JSON input/output for different AI/ML frameworks, libraries, compilers,
models and datasets, connect them into unified workflows instead
of hardwired scripts, and make them portable~\cite{ck-portable-workflows}
using automatic software detection plugins~\cite{ck-soft-plugins} and
meta-packages~\cite{ck-meta-packages}.
It also helps to make research more reliable and reproducible
by decomposing complex computational systems 
into reusable, portable, customizable, and non-virtualized CK components.
Finally, the CK concept is to be non-intrusive and complement,
abstract and interconnect all existing tools including MLFlow, SageMaker,
Kedro, Spack, EasyBuild, MLPerf, Docker, and Kubernetes while making them
more adaptive and system aware rather than replacing or rewriting them.

My long-term objective is to provide a common research infrastructure with different levels of abstraction 
that can bridge the gap between researchers and practitioners 
and help them to collaboratively co-design complex computational systems 
that can be immediately used in production as shown in Figure~\ref{fig:timeline2}.
Scientists can then work with a higher-level abstraction of such a system while allowing engineers
to continue improving the lower-level abstractions for evolving software and hardware without breaking the system.

Furthermore, the unified interfaces and meta descriptions of all CK components and workflows make it possible
to better understand what is happening inside complex and "black box" computational systems, 
integrate them with production and legacy systems, use them inside Docker and Kubernetes, 
share them along with published papers, and apply the DevOps methodology and agile principles in scientific research.


\section{Collective Knowledge platform}

\begin{figure*}[ht]
  \centering
  \includegraphics[width=1.0\textwidth]{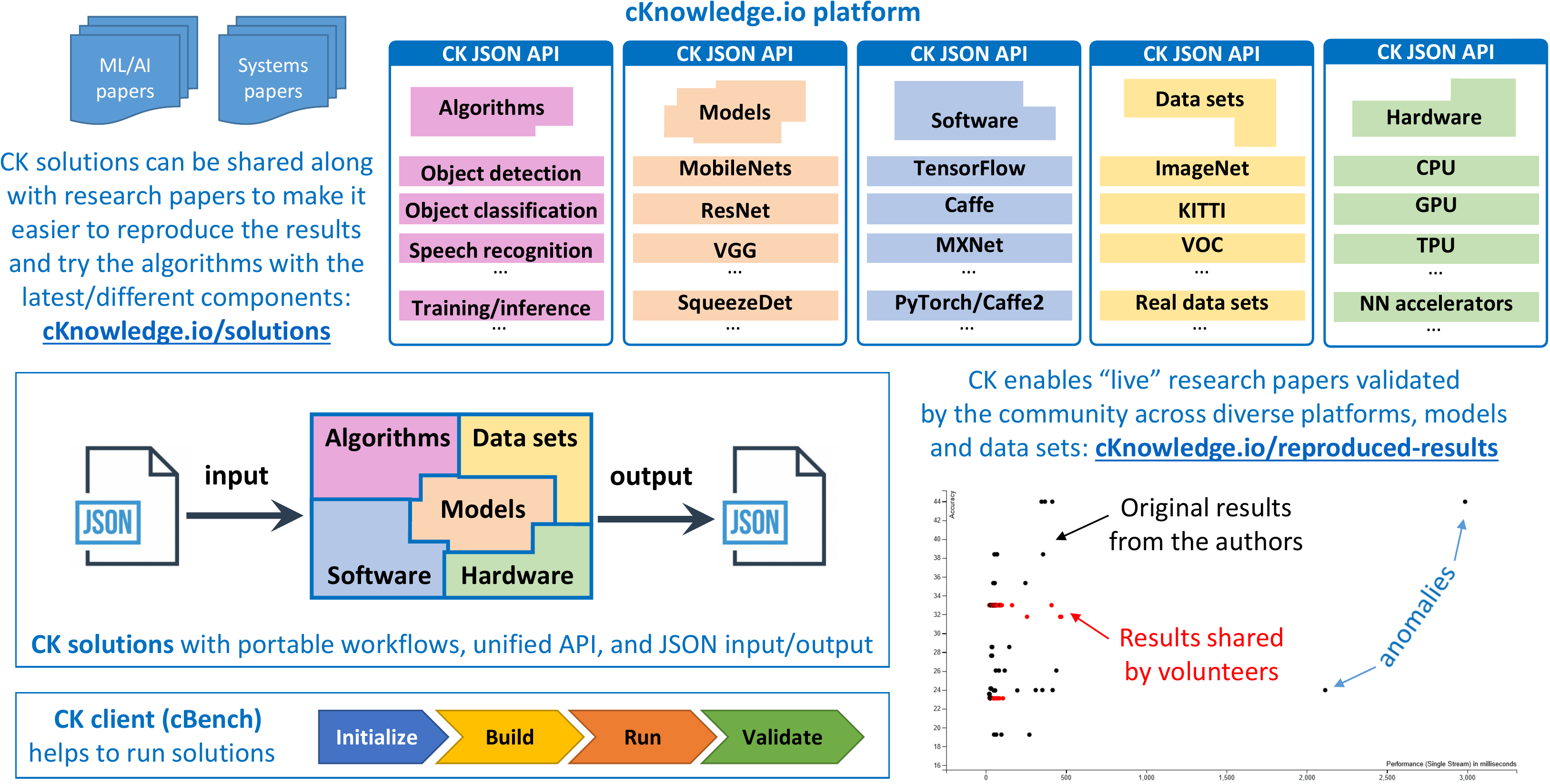}
  \caption{cKnowledge.io: a prototype of an open platform to share and reuse all the basic blocks and APIs 
   needed to co-design efficient and self-optimizing computational systems,
   enable live papers validated by the community, and keep track of the state-of-the-art machine learning and systems research 
   with the help of portable workflows and reproducible crowd-benchmarking.}
  \label{fig:ck2}
\end{figure*}

During the past 4 years, CK has been validated in different academic and industrial projects 
as a portable and modular workflow framework.
CK helped to enable reproducible experiments,
optimize software and hardware stacks for emerging AI, ML and quantum workloads,
bridge the gap between high-level ML operations and systems, and support MLOps~\cite{ck-projects}.
The authors of 18 research papers used CK to share their research artifacts 
and workflows at different ML\&systems conferences~\cite{ck-ae}.

While CK helped to automate benchmarking, optimization and co-design 
of complex computational systems and make it easier to deploy them in 
production~\cite{GM} I also noticed three major limitations:
  
\begin{itemize} 

 \item The distributed nature of the CK technology, the lack of
 a centralized place to keep all CK APIs, workflows and components,
 and the lack of convenient GUIs made it very challenging to keep track of all 
 contributions from the community, add new components, assemble workflows, automatically test
 them across diverse platforms, and connect them with legacy systems.

 \item The concept of backward compatibility of CK APIs and the lack
 of versioning similar to Java made it challenging to keep stable and
 bug-free workflows in real life - any bug in a reusable CK component 
 from one GitHub project could easily break dependent workflows 
 in another GitHub project.

 \item CK command-line interface was too low-level and not very user friendly.

\end{itemize} 

The feedback from CK users motivated me to start developing \href{https://cKnowledge.io}{cKnowledge.io} 
(Figure~\ref{fig:ck2}) - an open web-based platform to aggregate, version
and test all CK components, APIs, and portable CK workflows.
This is necessary to support collaborative and reproducible ML\&systems research and deploy
ML models in production across diverse systems, data sets and environments from IoT to data centers.
The CK platform is inspired by GitHub and PyPI: I see it as
a collaborative platform to share reusable research APIs, 
portable workflows, reproducible solutions, and associated reproducible results.
It also includes the open-source CK client~\cite{ck-client} to provide a common API to initialize, 
build, run and validate different research projects based on a simple JSON manifest describing all CK dependencies 
and installation/execution/validation recipes for different tasks and target platforms.
Such a platform can be used to keep track of reproducible and reusable state-of-the-art AI, ML
and systems research by connecting portable workflows and reusable artifacts with live scoreboards 
to validate and compare experimental results during Artifact Evaluation 
at different conferences~\cite{reproduced-results}.

I believe that the combination of the CK framework and the CK platform 
can make it easier to implement and share portable workflows for research code
assembled from stable and versioned CK components with unified APIs.
Such modular workflows can help to keep track of all the information flow within such workflows, 
expose and modify all configuration and optimization parameters via simple JSON input files, 
combine public and private code and data, 
monitor system behavior, retarget research code and machine learning models 
to different platforms from IoT to cloud, 
use them inside containers, integrate them with legacy systems, 
reproduce results, and generate reproducible papers with live scoreboards.

\section{Collective Knowledge use cases}

\begin{figure*}[ht]
  \centering
  \includegraphics[width=1.0\textwidth]{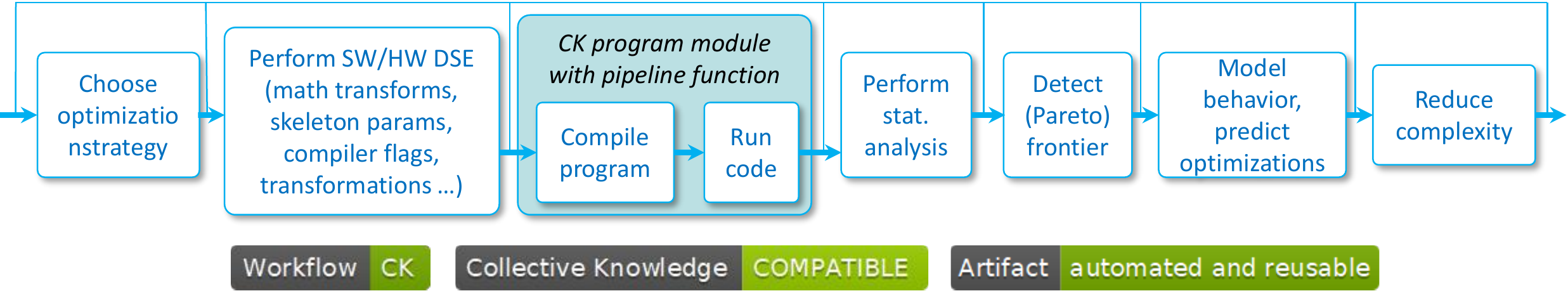}
  \caption{Reusable benchmarking and autotuning pipeline assembled from portable CK components with unified APIs.}
  \label{fig:pipeline}
\end{figure*}

As the first practical use case, I decided to convert all artifacts, workflows and automation tasks
from my past research related to self-learning and self-optimizing computer systems 
into reusable CK components.
I shared them with the community in CK-compatible Git repositories~\cite{ck-repos}
and started reproducing my past research results with new software, hardware, data sets 
and deep learning models~\cite{reproduced-results}.
I also implemented a customizable and portable benchmarking and autotuning pipeline (workflow)
that could perform software/hardware co-design in a unified way across different programs, data sets, frameworks,
compilers and platforms as shown in Figure~\ref{fig:pipeline}.

Such a pipeline helped to gradually expose different design choices and optimization parameters 
from all sub-components (models, frameworks, compilers, run-time systems, hardware) via unified CK APIs
and enable the whole system autotuning.
It also helped to keep track of all information passed between sub-components in complex
computational systems to ensure the reproducibility of results while finding the most efficient 
configuration on a Pareto frontier in terms of speed, accuracy, energy and other characteristics 
also exposed via unified CK APIs. 

I then decided to validate the CK concept of reusability by using the same pipeline
in another collaborative project with the Raspberry Pi foundation.
The practical task was to crowdsource compiler autotuning across multiple Raspberry Pi devices
to improve their performance.
CK helped to automate experiments, collect performance numbers on live CK scoreboards, 
and plug in CK components with various machine learning and predictive analytics techniques 
including decision trees, nearest neighbor classifiers, support vector machines (SVM) and deep learning
to automatically learn the most efficient optimizations~\cite{cm:29db2248aba45e59:c4b24bff57f4ad07}.
It also demonstrated the possibility to reduce the growing technology transfer gap between academia and industry
by reusing portable workflows that can adapt to evolving systems and can be integrated with existing and legacy projects.
For example, the same pipeline was successfully reused by General Motors to collaboratively benchmark and optimizing deep learning implementations~\cite{GM}
and by Amazon to enable scaling of deep learning on AWS using C5 instances with MXNet, TensorFlow, and BigDL from the edge to the cloud~\cite{ck-amazon}.
Finally, my CK autotuning pipeline was reused and extended by dividiti 
to make it easier to prepare, submit and reproduce MLPerf inference benchmark 
results~\cite{mlperf-inference-results}.

\section{Collective Knowledge demo: automating, sharing and reproducing MLPerf inference benchmarks}

I prepared a live and interactive demo of the CK solution that automates 
the MLPerf inference benchmark, connects it with the live CK dashboard and
crowdsource benchmarking across diverse platforms provided by volunteers
similar to SETI@home: \href{https://cKnowledge.io/demo}{cKnowledge.io/demo}.
This demo shows how to use CK APIs to automatically build, run 
and validate object detection based on SSD-Mobilenet, TensorFlow and COCO dataset 
across Raspberry Pi computers, Android phones, laptops, desktops, and data centers.
This solution is based on a simple JSON file describing the following tasks
and their dependencies on CK components:
\begin{itemize} 
 \item prepare a Python virtual environment (can be skipped for the native installation),
 \item download and install the Coco dataset (50 or 5000 images),
 \item detect C++ compilers or Python interpreters needed for object detection,
 \item install Tensorflow framework with a specified version for a given target machine,
 \item download and install the SSD-MobileNet model compatible with selected Tensorflow,
 \item manage the installation of all other dependencies and libraries,
 \item compile object detection for a given machine and prepare pre/post-processing scripts.
\end{itemize}

This solution was published on the \href{https://cKnowledge.io}{cKnowledge.io} platform 
using the open-source CK client~\cite{ck-client} to help users to participate in crowd-benchmarking 
using their own machines as follows:
\begin{enumerate}
    \item install CK client from PyPi using:
    
    \textit{pip install cbench}
    
    \item download and install the solution on a given machine (example for Linux): 
    
    \textit{cb init demo-obj-detection-coco-tf-cpu-benchmark-linux-portable-workflows}
    
    \item run the solution on a given machine:
    
    \textit{cb benchmark demo-obj-detection-coco-tf-cpu-benchmark-linux-portable-workflows}
\end{enumerate}

The users can then see their measurements (speed, latency, accuracy and other exposed characteristics)
and compare them against the official MLPerf results or with the results shared by other users
with the help of the live CK dashboard associated with this solution: 
\href{https://cKnowledge.io/result/sota-mlperf-object-detection-v0.5-crowd-benchmarking}{cKnowledge.io/result/sota-mlperf-object-detection-v0.5-crowd-benchmarking}.

After validating this solution on a given platform, the users can also clone it and update the JSON description
to retarget this benchmark to other devices and operating systems such as macOS, Windows with
Docker, Android phones, servers with CUDA-enabled GPUs, and so on.

Finally, the users can integrate such ML solutions with production systems with the help of unified CK APIs
as demonstrated by connecting above CK solution for object detection with the webcam in the browser:
\href{https://cKnowledge.io/solution/demo-obj-detection-coco-tf-cpu-webcam-linux-azure}{cKnowledge.io/solution/demo-obj-detection-coco-tf-cpu-webcam-linux-azure}.

\section{Conclusions and future work}

My very first research project to prototype semiconductor neural network stalled in the late 90s
because it took me way too long to build all the infrastructure from scratch 
to generate data sets, train neural networks, prepare and optimize hardware,
run all experiments and optimize the NN implementation.
Since then, I have always been looking for solutions to enable more efficient computer systems 
and accelerate ML\&systems research.

I have developed the CK framework and cKnowledge.io platform to bring DevOps, MLOps, reusability and agile principles to ML\&systems research,
and connect researchers and practitioners to co-design more reliable, reproducible and efficient computational systems 
that can adapt to continuously changing software, hardware, models, and data sets.
I hope that CK will help to share and reuse best practices and pack research techniques and artifacts along with research papers 
to make it easier to reproduce results and deploy them in production.
I also want to enable "live" research papers by connecting CK workflows with live dashboards 
to let the community reproduce results, detect unexpected behavior, and collaboratively fix problems 
in shared workflows and components~\cite{cm:29db2248aba45e59:c4b24bff57f4ad07}.

However, CK is still a proof-of-concept and there is a lot to be improved.
For example, I would like to make it more user friendly, standardize APIs and JSON meta descriptions 
of all CK components and workflows, and develop a simple GUI to share CK components,
assemble workflows, run experiments, and compare research techniques similar to LEGO.
My dream is to use CK to build a virtual world (playground) where researchers and practitioners 
assemble AI, ML and other novel algorithms similar to live species that can continue to evolve, self-optimize 
and compete with each other across devices and data provided by volunteers.
At the same time, the winning solutions with the best trade-off in terms of speed, latency, accuracy, energy, size, and costs 
can be simplify picked from the Pareto frontier at any time and immediately deployed in production 
thus accelerating AI, ML and systems research and making AI practical.

\section*{Software and Data}

All code and data can be found at \href{https://cKnowledge.io}{cKnowledge.io} under permissive license.

\section*{Acknowledgements}

I would like to thank 
Unmesh D. Bordoloi,
Nikolay Chunosov,
Marco Cianfriglia,
Cody Coleman,
Achi Dosanjh,
Debojyoti Dutta,
Nicolas Essayan,
Todd Gamblin,
Leo Gordon,
Wayne Graves,
Herve Guillou,
Michael Heroux,
James Hetherignton,
Ivo Jimenez,
Gaurav Kaul,
Sergey Kolesnikov,
Dan Laney,
Anton Lokhmotov,
Peter Mattson,
Thierry Moreau,
Dewey Murdick,
Cedric Nugteren,
Bhavesh Patel,
Gennady Pekhimenko,
Massimiliano Picone,
Ed Plowman,
Vijay Janapa Reddi,
Alka Roy,
Michel Steuwer,
Victoria Stodden,
Robert Stojnic,
Stuart Taylor,
Eben Upton,
Flavio Vella,
Boris Veytsman,
Alex Wade,
Matei Zaharia,
Alex Zhigarev,

and many other ACM, IEEE, NeurIPS and MLPerf colleagues 
for interesting discussions, practical use cases and useful feedback.

\bibliographystyle{plain}
\bibliography{ck}

\begin{thebibliography}{10}

\bibitem{sagemaker}
Amazon sagemaker: fully managed service to build, train, and deploy machine
  learning models quickly.
\newblock \url{https://aws.amazon.com/sagemaker}.

\bibitem{ck-ae-appendix}
{Artifact Appendix and reproducibility checklist to unify and automate the
  validation of results at systems and machine learning conferences}.
\newblock \url{https://ctuning.org/ae/submission_extra.html}.

\bibitem{AE}
{Artifact Evaluation: reproducing experimental results from published research
  papers at systems and machine learning conferences (the common methodology,
  automation tools, artifact appendix and reproducibility checklist)}.
\newblock \url{https://cTuning.org/ae}.

\bibitem{ck-actions}
{Automation actions with a unified API and JSON IO to share best practices and
  introduce DevOps principles to scientific research}.
\newblock \url{https://cKnowledge.io/actions}.

\bibitem{ck-repos}
{CK-compatible GitHub, GitLab and BitBucket repositories with CK APIs,
  components and workflows}.
\newblock \url{https://cKnowledge.io/repos}.

\bibitem{GM}
{CK use case from General Motors: collaboratively Benchmarking and Optimizing
  Deep Learning Implementations}.
\newblock \url{https://youtu.be/1ldgVZ64hEI}.

\bibitem{ck-projects}
{Collective Knowledge real-world use cases}.
\newblock \url{https://cKnowledge.org/partners}.

\bibitem{ck-client}
{Cross-platform CK client to unify preparation, execution and validation of
  research techniques shared along with research papers}.
\newblock \url{https://github.com/ctuning/cbench}.

\bibitem{reproduced-results}
{Live scoreboards with reproduced results from research papers at machine
  learning and systems conferences}.
\newblock \url{https://cKnowledge.io/reproduced-results}.

\bibitem{mlperf-inference-results}
{MLperf inference benchmark results}.
\newblock \url{https://mlperf.org/inference-results}.

\bibitem{ck-portable-workflows}
{Portable and reusable CK workflows}.
\newblock \url{https://cKnowledge.io/programs}.

\bibitem{ae-reproduced-papers}
{Reproduced papers with ACM badges based on the cTuning evaluation
  methodology}.
\newblock \url{https://cKnowledge.io/reproduced-papers}.

\bibitem{ck-modules}
{Shared and versioned CK modules}.
\newblock \url{https://cKnowledge.io/modules}.

\bibitem{ck-meta-packages}
{Shared CK meta packages (code, data, models) with automated installation
  across diverse platforms}.
\newblock \url{https://cKnowledge.io/packages}.

\bibitem{ck-soft-plugins}
{Shared CK plugins to detect software (models, frameworks, data sets, scripts)
  on a user machine}.
\newblock \url{https://cKnowledge.io/soft}.

\bibitem{ck-ae}
{Artifact Evaluation: reproducing experimental results from systems and machine
  learning conferences}.
\newblock \url{https://cTuning.org/ae}, 2015--current.

\bibitem{deep500}
T.~Ben-Nun, M.~Besta, S.~Huber, A.~N. Ziogas, D.~Peter, and T.~Hoefler.
\newblock {A Modular Benchmarking Infrastructure for High-Performance and
  Reproducible Deep Learning}.
\newblock IEEE, May 2019.
\newblock The 33rd IEEE International Parallel \& Distributed Processing
  Symposium (IPDPS'19).

\bibitem{cfkz2016}
Bruce~R. Childers, Grigori Fursin, Shriram Krishnamurthi, and Andreas Zeller.
\newblock {Artifact Evaluation for Publications (Dagstuhl Perspectives Workshop
  15452)}.
\newblock {\em Dagstuhl Reports}, 5(11):29--35, 2016.

\bibitem{Fur2009}
Grigori Fursin.
\newblock {Collective Tuning Initiative: automating and accelerating
  development and optimization of computing systems}.
\newblock In {\em Proceedings of the GCC Developers' Summit}, June 2009.

\bibitem{ck2}
Grigori Fursin.
\newblock {CK: an open-source and cross-platform framework to decompose complex
  computational systems and research projects into portable, customizable and
  reusable components and workflows with Python APIs, JSON input/output files
  and an integrated meta package manager}.
\newblock \url{https://github.com/ctuning/ck}, 2015--present.

\bibitem{grigori_fursin_2017_2544204}
Grigori Fursin.
\newblock {CNRS presentation: enabling open and reproducible research at
  computer systems conferences: the good, the bad and the ugly}.
\newblock https://doi.org/10.5281/zenodo.2544204, March 2017.

\bibitem{DBLP:journals/corr/FursinD14}
Grigori Fursin and Christophe Dubach.
\newblock Community-driven reviewing and validation of publications.
\newblock {\em CoRR}, abs/1406.4020, 2014.

\bibitem{ck-date16}
Grigori Fursin, Anton Lokhmotov, and Ed~Plowman.
\newblock {Collective Knowledge}: towards {R\&D} sustainability.
\newblock In {\em Proceedings of the Conference on Design, Automation and Test
  in Europe (DATE'16)}, March 2016.

\bibitem{cm:29db2248aba45e59:c4b24bff57f4ad07}
Grigori {Fursin}, Anton {Lokhmotov}, Dmitry {Savenko}, and Eben {Upton}.
\newblock {A Collective Knowledge workflow for collaborative research into
  multi-objective autotuning and machine learning techniques}.
\newblock January 2018.

\bibitem{spack}
T.~{Gamblin}, M.~{LeGendre}, M.~R. {Collette}, G.~L. {Lee}, A.~{Moody}, B.~R.
  {de Supinski}, and S.~{Futral}.
\newblock The spack package manager: bringing order to hpc software chaos.
\newblock In {\em SC '15: Proceedings of the International Conference for High
  Performance Computing, Networking, Storage and Analysis}, pages 1--12, Nov
  2015.

\bibitem{easybuild}
Kenneth Hoste, Jens Timmerman, Andy Georges, and Stijn Weirdt.
\newblock Easybuild: Building software with ease.
\newblock pages 572--582, 11 2012.

\bibitem{ck-amazon}
Gaurav Kaul, Suneel Marthi, and Grigori Fursin.
\newblock {Scaling deep learning on AWS using C5 instances with MXNet,
  TensorFlow, and BigDL: From the edge to the cloud}.
\newblock
  https://conferences.oreilly.com/artificial-intelligence/ai-eu-2018/public/schedule/detail/71549,
  2018.

\bibitem{li2019acrossstack}
Cheng Li, Abdul Dakkak, Jinjun Xiong, Wei Wei, Lingjie Xu, and Wen mei Hwu.
\newblock Across-stack profiling and characterization of machine learning
  models on gpus.
\newblock https://arxiv.org/abs/1908.06869, 2019.

\bibitem{mdhs2009}
Todd Mytkowicz, Amer Diwan, Matthias Hauswirth, and Peter~F. Sweeney.
\newblock Producing wrong data without doing anything obviously wrong!
\newblock In {\em Proceedings of the 14th International Conference on
  Architectural Support for Programming Languages and Operating Systems}, page
  265–276, 2009.

\bibitem{kedro}
QuantumBlack.
\newblock {Kedro}: the open source library for production-ready machine
  learning code.
\newblock \url{https://github.com/quantumblacklabs/kedro}, 2019.

\bibitem{reddi2019mlperf}
Vijay~Janapa Reddi, Christine Cheng, David Kanter, Peter Mattson, Guenther
  Schmuelling, Carole-Jean Wu, Brian Anderson, Maximilien Breughe, Mark
  Charlebois, William Chou, Ramesh Chukka, Cody Coleman, Sam Davis, Pan Deng,
  Greg Diamos, Jared Duke, Dave Fick, J.~Scott Gardner, Itay Hubara, Sachin
  Idgunji, Thomas~B. Jablin, Jeff Jiao, Tom~St. John, Pankaj Kanwar, David Lee,
  Jeffery Liao, Anton Lokhmotov, Francisco Massa, Peng Meng, Paulius
  Micikevicius, Colin Osborne, Gennady Pekhimenko, Arun Tejusve~Raghunath
  Rajan, Dilip Sequeira, Ashish Sirasao, Fei Sun, Hanlin Tang, Michael Thomson,
  Frank Wei, Ephrem Wu, Lingjie Xu, Koichi Yamada, Bing Yu, George Yuan, Aaron
  Zhong, Peizhao Zhang, and Yuchen Zhou.
\newblock Mlperf inference benchmark.
\newblock https://arxiv.org/abs/1911.02549, 2019.

\bibitem{Zaharia2018AcceleratingTM}
Matei Zaharia, Andrew Chen, Aaron Davidson, Ali Ghodsi, Sue~Ann Hong, Andy
  Konwinski, Siddharth Murching, Tomas Nykodym, Paul Ogilvie, Mani Parkhe, Fen
  Xie, and Corey Zumar.
\newblock Accelerating the machine learning lifecycle with mlflow.
\newblock {\em IEEE Data Eng. Bull.}, 41:39--45, 2018.

\end{thebibliography}

\end{document}